\documentclass[conference]{IEEEtran}
\IEEEoverridecommandlockouts
\usepackage{cite}
\usepackage{amsmath,amssymb,amsfonts}
\usepackage{algorithmic}
\usepackage{graphicx}
\usepackage{textcomp}
\usepackage[dvipsnames]{xcolor}
\usepackage{booktabs}
\usepackage{colortbl}
\usepackage{url}
\usepackage{multicol}
\usepackage{multirow}
\usepackage{soul}
\def\BibTeX{{\rm B\kern-.05em{\sc i\kern-.025em b}\kern-.08em
    T\kern-.1667em\lower.7ex\hbox{E}\kern-.125emX}}

\usepackage{fancyhdr}
\usepackage{tikz}
\newcommand\copyrighttext{%
    © 2024 IEEE.  Personal use of this material is permitted.  Permission from IEEE must be obtained for all other uses, in any current or future media, including reprinting/republishing this material for advertising or promotional purposes, creating new collective works, for resale or redistribution to servers or lists, or reuse of any copyrighted component of this work in other works.}
\newcommand\copyrightnotice{%
\begin{tikzpicture}[remember picture,overlay]
\node[anchor=south,yshift=65pt] at (current page.south) {\fbox{\parbox{\dimexpr\textwidth-\fboxsep-\fboxrule\relax}{\copyrighttext}}};
\end{tikzpicture}%
}

\begin{document}
\copyrightnotice

\title{
Guardians of Discourse: Evaluating LLMs on Multilingual Offensive Language Detection
}

\author{\IEEEauthorblockN{Jianfei He$^1$, Lilin Wang$^1$, Jiaying Wang$^3$, Zhenyu Liu$^1$, Hongbin Na$^4$, Zimu Wang$^2$, Wei Wang$^2$, Qi Chen$^{1,}$\IEEEauthorrefmark{2}\thanks{\IEEEauthorrefmark{2}Corresponding author.}}
\IEEEauthorblockA{
\textit{$^1$School of AI and Advanced Computing, Xi'an Jiaotong-Liverpool University, Suzhou, China} \\
\textit{$^2$School of Advanced Technology, Xi'an Jiaotong-Liverpool University, Suzhou, China} \\
\textit{$^3$College of Computer Science and Technology, Zhejiang University of Technology, Hangzhou, China} \\
\textit{$^4$Australian Artificial Intelligence Institute, University of Technology Sydney, Sydney, Australia} \\
Jianfei.He20@student.xjtlu.edu.cn, Qi.Chen02@xjtlu.edu.cn}
}

\maketitle

\begin{abstract}
Identifying offensive language is essential for maintaining safety and sustainability in the social media era. Though large language models (LLMs) have demonstrated encouraging potential in social media analytics, they lack thorough evaluation when in offensive language detection, particularly in multilingual environments. We for the first time evaluate multilingual offensive language detection of LLMs in three languages: English, Spanish, and German with three LLMs, GPT-3.5, Flan-T5, and Mistral, in both monolingual and multilingual settings. We further examine the impact of different prompt languages and augmented translation data for the task in non-English contexts. Furthermore, we discuss the impact of the inherent bias in LLMs and the datasets in the mispredictions related to sensitive topics.
\end{abstract}


\textcolor{red}{\textit{\textbf{Warning}: This paper contains content that may be offensive or upsetting.}}

\section{Introduction}

\begin{figure}[t!]
    \centering
    \includegraphics[width=0.94\linewidth]{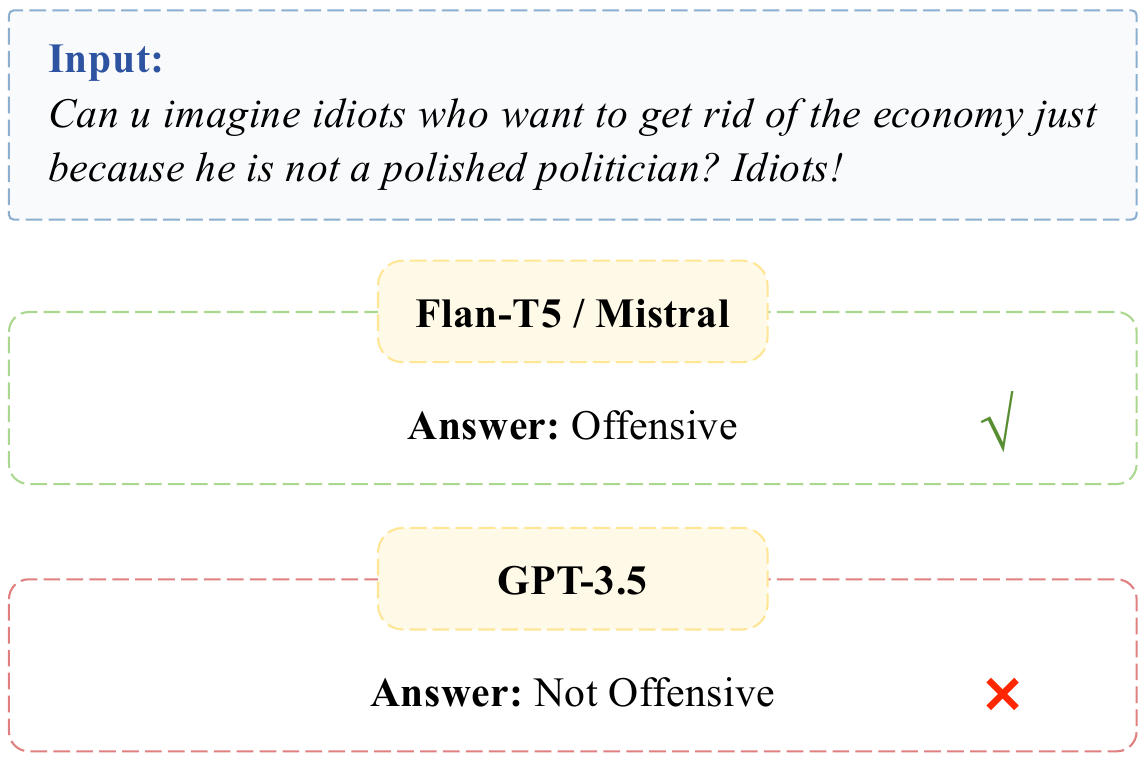}
    \caption{Offensive language detection results for an example tweet obtained from three LLMs: Flan-T5, Mistral, and GPT-3.5.}
    \label{fig:example}
    \vspace{-2mm}
\end{figure}

The issue of online offensive language is growing increasingly and attracting attention from policymakers, civil society organisations, and the public \cite{stockmann2023social}. With their anonymity and global reach, social media platforms have significantly contributed to the spread of offensive content, which negatively impacts various individuals and communities \cite{pradhan2020review}. Given the proliferation of offensive language on platforms like X (Twitter), numerous techniques, particularly based on pre-trained language models (PLMs), have been proposed to identify such content, particularly in English contexts.
However, existing research on multilingual offensive language detection is still in its infancy \cite{mnassri2024survey}. Recently, large language models (LLMs) have shown promising performance in handling large-scale data and complex language tasks \cite{10.1145/3543873.3587368,peng2023does,10580402,YiningLarge2024,wang-etal-2024-knowledge}, showing their great potential in precisely identifying offensive content in social media (Figure \ref{fig:example}). Nevertheless, no existing work has extensively evaluated the ability of LLMs in terms of multilingual offensive detection.

In this paper, we conduct the first comprehensive evaluation to demonstrate LLMs’ ability in multilingual offensive language detection. We select three datasets in diverse languages: OLID + SOLID (English) \cite{rosenthal2020solid}, OffendES (Spanish) \cite{Arco2021OffendESAN}, and GermEval 2018 (German) \cite{Wiegand2018OverviewOT} and three LLMs: GPT-3.5, Flan-T5 \cite{chung2022scaling}, and Mistral \cite{jiang2023mistral}, and compare the results with the previous state-of-the-art (SOTA) methods. As shown in Figure \ref{fig:framework}, our evaluation procedures are as follows:

\begin{figure*}[t!]
    \centering 
    \includegraphics[width=0.9\linewidth]{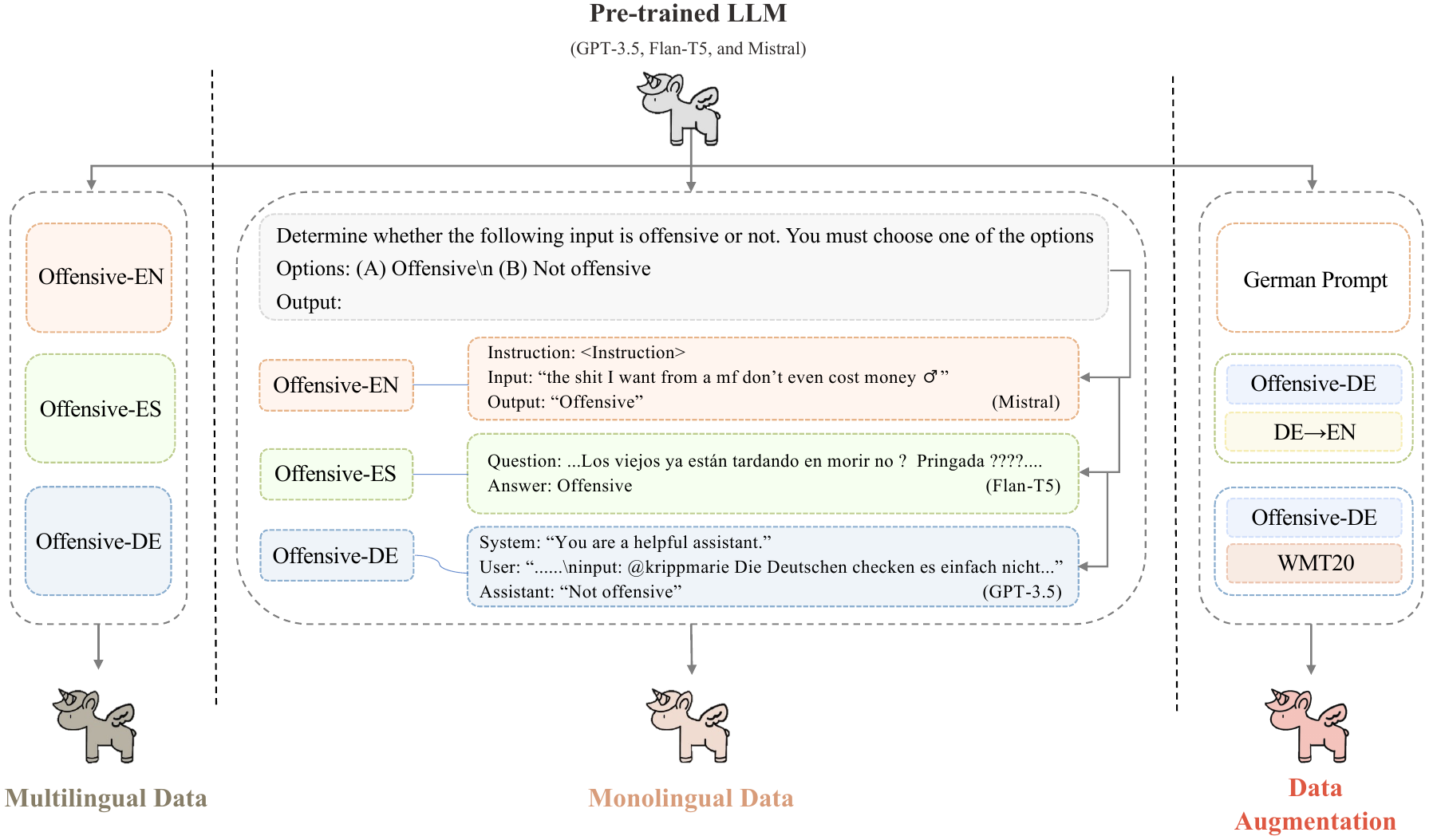}
    \vspace{-2mm}
    \caption{Evaluation pipeline of LLMs in multilingual offensive language detection.}
    \label{fig:framework}
    \vspace{-2mm}
\end{figure*}

First, we assess the capabilities of LLMs in detecting offensive content within a \textit{monolingual} environment by fine-tuning the experimented models. Afterwards, we explore the impacts of \textit{multilingual} data on the task by fine-tuning LLMs with a mixture of multilingual data.
Besides, following the previous work that has demonstrated the effectiveness of \textit{data augmentation} \cite{conneau2019cross,siddhant2020evaluating} and the efficacy of translation data in non-English tasks \cite{zhu2023extrapolating}, we select German as our experimented language and conduct the following additional investigations: 1) \textit{Native language prompts}: We construct the instruction with the native language (German) rather than English to test the effectiveness when the prompt matches the dataset language; and 2) \textit{Incorporating translation data}: We incorporate two types of translation data, including the translation of the English data and the data from WMT20 \cite{barrault-etal-2020-findings}, to assess whether LLMs can learn from translation data in non-English offensive language detection.

The key findings and insights of this work are as follows:

\begin{itemize}
    \item LLMs can achieve comparable or better performance in multilingual language detection. GPT-3.5 and Mistral have robust performance in all languages; however, Flan-T5 can only master English offensive language detection.
    \item Multilingual fine-tuning LLMs enhances models' offensive language detection ability. Particularly, Flan-T5 benefits the most among all experimented models.
    \item Utilising prompts in native languages enhance model comprehension in non-English contexts; however,  incorporating translation data cannot improve performance.
    \item The inherent biases in LLMs and the dataset in race, sexual orientation, and notably, genders, result in the incorrect prediction in sensitive topics.
\end{itemize}

\section{Related Work}

The detection of offensive language has emerged as a significant research area, particularly within the context of online communication. We review previous research divided into traditional methods and LLM-based approaches.

\paragraph{Traditional Methods} Previous research has made significant contributions to improving offensive language detection capabilities. For example, Vashistha and Zubiaga \cite{vashistha2020online} and Aluru et al. \cite{aluru2020deep} employ Convolutional Neural Network (CNN) and Long Short-Term Memory (LSTM) to capture spatial and temporal text features, respectively. Some other work \cite{ibrohim2019translated, aluru2020deep, jiang2021cross} utilise machine translation methods, thereby expanding the scope of multilingual offensive language detection. Furthermore, Rizwan et al. \cite{rizwan2020automatic} and Awal et al. \cite{awal2023modelagnostic} propose meta-learning strategies to address data scarcity issues.

The aforementioned methods demonstrate that enhancing model adaptability and applying data augmentation can effectively improve performance. Motivated by these work, we investigate the efficacy of the translation-based data augmentation method for German offensive language detection, including leveraging German prompts or performing a joint training with the translation of the English data and the data from WMT20 with the original dataset.

\paragraph{LLM-based Approaches} Motivated by their promising performance, LLMs have also been adopted in this task. Zampieri et al. \cite{zampieri2023offenseval} evaluates the performance of six LLMs and two PLMs and demonstrates the competitiveness of Flan-T5. Yang et al. \cite{yang2023hare} proposes HARE, a step-by-step reasoning method. Casula et al. \cite{casula2023generation} leverages GPT-2 and investigates the efficacy of multiple data augmentation methods. However, the ability of LLM to detect offensive text in a multilingual context has yet to be fully evaluated. We conduct the first evaluation to demonstrate LLMs' ability in multilingual offensive language detection and the impact of different data augmentation strategies for non-English scenarios.

\section{Evaluation Settings}

\subsection{Datasets and Evaluation Metrics}

We conducted experiments on three widely used offensive language detection datasets:
\begin{enumerate}
    \item \textbf{OLID} (\textit{Offensive Language Identification Dataset}) + \textbf{SOLID} (\textit{Semi-supervised OLID}) \cite{rosenthal2020solid}, which contains $3$ tasks, $7$ labels and around $12$ billion tweets;
    \item \textbf{OffendES} \cite{Arco2021OffendESAN}, which contains $2$ tasks, $5$ labels and $47,128$ comments; and
    \item \textbf{GermEval} (\textit{GermEval 2018}, \cite{Wiegand2018OverviewOT}), which contains $2$ tasks, $6$ labels and $8,541$ tweets.
\end{enumerate}

We selected macro precision, macro recall, and macro F1-score as evaluation metrics to be consistent with the previous work \cite{wiedemann2020uhh,Molero2023OffensiveLD,montani2018tuwienkbs}, and the accuracy of offensive samples (OFF), non-offensive samples (NON), and the entire dataset (ALL).

\subsection{Data Pre-processing}

Following the previous work in the benckmarking on LLMs \cite{peng2023does,na2024rethinking}, we sampled subsets from the datasets for efficient training and evaluation. We sampled the English and the Spanish datasets of $5,000$ samples in a ratio $1:1$ between ``\textit{Offensive}'' and ``\textit{Not offensive}'', and we leveraged the whole $5,009$ training samples for the German dataset. Due to the variety of labels, we adopted the same method as described in \cite{Arco2021OffendESAN} to map the labels into two for binary classification. 

\begin{table}[t!]
\centering
\small
\caption{Instructions and Examples for Evaluation}
\resizebox{\linewidth}{!}{
\begin{tabular}{l}
\toprule
\rowcolor{Salmon!10}
\multicolumn{1}{c}{{\textbf{English Instruction for Offensive Language Detection}}}                                   \\ \midrule
\rowcolor{Salmon!10}
\textbf{{[}Instruction{]}}                                                                                    \\
\rowcolor{Salmon!10}
Determine whether the following input is offensive or not. \\

\rowcolor{Salmon!10}
You must choose one of the options. \\

\rowcolor{Salmon!10}
Options: (A) Offensive (B) Not offensive              \\                                                         
\rowcolor{Salmon!10}
Output:                                                                                                       \\
\rowcolor{Salmon!10}
\textbf{{[}Input{]}}                                                                                          \\
\rowcolor{Salmon!10}
Liebe Corinna, [...] Wärst du begeisterbar? \\
\rowcolor{Salmon!10}
\textbf{{[}Output{]}}                                                                                         \\
\rowcolor{Salmon!10}
Not offensive                                                                                                 \\ \midrule
\rowcolor{CornflowerBlue!10}
\multicolumn{1}{c}{{\textbf{German Instruction for Offensive Language Detection}}}                                    \\ \midrule
\rowcolor{CornflowerBlue!10}
\textbf{{[}Instruction{]}}                                                                                    \\
\rowcolor{CornflowerBlue!10}
Stellen Sie fest, ob die folgende Eingabe anstößig ist oder nicht. \\
\rowcolor{CornflowerBlue!10}
Sie müssen eine der Optionen auswählen.    \\
\rowcolor{CornflowerBlue!10}
Optionen: (A) Beleidigend (B) Nicht beleidigend                                                               \\
\rowcolor{CornflowerBlue!10}
Ausgabe:                                                                                                      \\
\rowcolor{CornflowerBlue!10}
\textbf{{[}Input{]}}                                                                                          \\
\rowcolor{CornflowerBlue!10}
Liebe Corinna, [...] Wärst du begeisterbar? \\
\rowcolor{CornflowerBlue!10}
\textbf{{[}Output{]}}                                                                                         \\
\rowcolor{CornflowerBlue!10}
Nicht beleidigend                                                                                             \\ \bottomrule
\end{tabular}
}
\vspace{-2mm}
\label{tab:prompt}
\end{table}

\subsection{Prompt Design}

Table \ref{tab:prompt} illustrates the prompts for our evaluation with examples, in which the English instruction was used in main experiments, and the German instruction was for additional analysis. Generally, the prompts consist of an instruction, an input text, and a placeholder for the output. We formulated the offensive language detection task as multiple choice questions that require LLMs to select one option from “\textit{Offensive}” and “\textit{Not offensive}” following \cite{yang2023hare}.

\subsection{Experimental Setup}

We conducted experiments on three LLMs: GPT-3.5, Flan-T5 \cite{chung2022scaling}, and Mistral \cite{jiang2023mistral}. For GPT-3.5, fine-tuned the \texttt{gpt-3.5-turbo-0125} checkpoint through OpenAI's official API\footnote{\url{https://platform.openai.com/}}, and we set the number of epoch as $5$. For Flan-T5 and Mistral, we selected \texttt{flan-t5-base} and \texttt{Mistral-} \texttt{7B-Instruct-v0.2}. During the fine-tuning process, we set the number of epochs as $5$, the learning rate as $5e-5$, and we adopted LLaMA-Factory \cite{zheng2024llamafactory} for fine-tuning with LoRA \cite{hu2022lora} when fine-tuning Mistral. The experiments were conducted on NVIDIA GeForce 4090 graphic cards with $26$ GPU hours.

\subsection{Baselines}

We compared the performance of LLMs on offensive language detection with the following SOTA methods:

\begin{itemize}
    \item \textit{English (OLID + SOLID)}: We compared the performance of LLMs with RoBERTa-Large implemented by \cite{wiedemann2020uhh}, which achieved the best performance in SemEval-2020.
    \item \textit{Spanish (OffendES)}: We referred to \cite{Molero2023OffensiveLD}, which incorporated Bag-of-Words representation to RoBERTuito, a PLM for social media text in Spanish.
    \item \textit{German (GermEval)}: We selected the TUWienKBS system \cite{montani2018tuwienkbs} that employed a stacked classifier architecture to combine information from multiple sources.
\end{itemize}

\section{Results and Analysis}

\begin{figure}[t!]
    \centering 
    \includegraphics[width=1\linewidth]{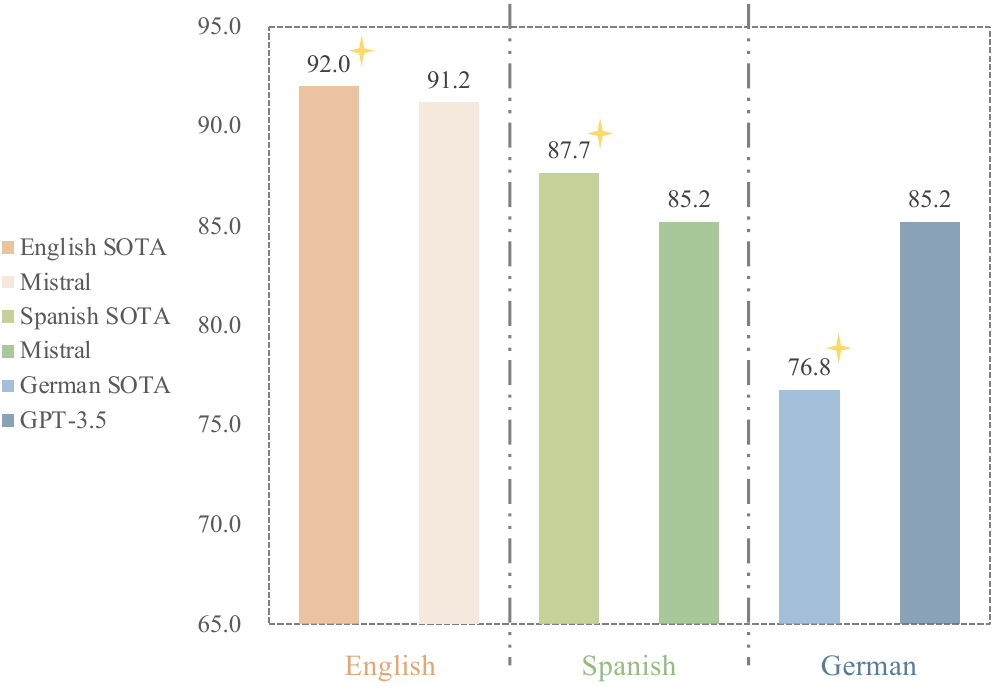}
    \caption{Comparison between previous SOTA methods and the best results with LLMs during the entire evaluation process.}
    \label{fig:bar}
    \vspace{-3mm}
\end{figure}

Figure \ref{fig:bar} illustrates the comparison between the previous SOTA methods and the best results during the entire evaluation process. Generally, LLMs achieved comparable or better performance in the task. The detailed results and findings will be explained in the following sections.

\subsection{Detection with Monolingual Data}
\paragraph{Model Performance Comparison}

Table \ref{tab:monolingual} illustrates the experimental results obtained from the LLMs fine-tuned from monolingual data. Generally, the performance of fine-tuned LLMs was comparable but could not surpass the SOTA methods except for the German dataset.
From the results, we had the following observations:

First, Mistral achieved the best performance in English and Spanish offensive language detection; however, GPT-3.5 outperformed Mistral in the German context. In the detection of English and Spanish offensive languages, LLMs achieved a lower precision, a higher recall, and a higher accuracy in offensive content, while a reversed case occurred in the German context. This phenomenon was attributed to the distribution of training data—the ratio between “\textit{Offensive}” and “\textit{Not offensive}” was $1:1$ in English and Spanish training data during data sampling, but the entire training set of the German dataset was used, whose ratio was around $1:2$.

Second, even though the number of parameters of Flan-T5 was much fewer than GPT-3.5 and Mistral, it achieved comparable performance in offensive language detection in the English context; however, it was much behind when detecting non-English contexts, such as Spanish and German, i.e. the F1-score of Flan-T5 was $8.8\%$ and $12.6\%$ fewer than GPT-3.5 in Spanish and German, respectively. \ul{This indicates that Flan-T5 masters English offensive language detection but is poor in the detection in non-English contexts.}

\begin{table}[t!]
\centering
\caption{Experimental Results of LLMs Trained by Monolingual Data}
\resizebox{\linewidth}{!}{
    \begin{tabular}{lccccccccc}
    \toprule
    \multirow{2}{*}{\textbf{Model}} & \multicolumn{9}{c}{\textbf{Language}} \\
    \cmidrule{2-10}
     & \multicolumn{3}{c}{English} & \multicolumn{3}{c}{Spanish} & \multicolumn{3}{c}{German} \\
    \midrule
     & \multicolumn{1}{c}{P} & \multicolumn{1}{c}{R} & \multicolumn{1}{c}{F1} &\multicolumn{1}{c}{P} & \multicolumn{1}{c}{R} & \multicolumn{1}{c}{F1}  & \multicolumn{1}{c}{P} & \multicolumn{1}{c}{R} & \multicolumn{1}{c}{F1} \\
    \cmidrule{2-10}
    SOTA & $90.1$ & $94.8$ & $\mathbf{92.0}$ & $90.1$ & $85.7$ & $\mathbf{87.7}$ & $77.4$ & $76.1$ & $76.8$ \\ 
    GPT-3.5 & $88.8$ & $94.4$ & $90.7$ & $80.9$ & $91.0$ & $84.2$ & $85.8$ & $83.6$ & $\mathbf{84.5}$ \\ 
    Flan-T5 & $88.5$ & $94.2$ & $90.5$ & $73.1$ & $80.7$ & $75.4$ & $73.1$ & $71.3$ & $71.9$ \\
    Mistral & $89.1$ & $94.2$ & $\underline{90.9}$ & $82.2$ & $87.7$ & $\underline{84.4}$ & $83.0$ & $80.0$ & $\underline{81.1}$ \\
    \midrule
     & \multicolumn{1}{c}{OFF} & \multicolumn{1}{c}{NOT} & \multicolumn{1}{c}{FULL} & \multicolumn{1}{c}{OFF} & \multicolumn{1}{c}{NOT} & \multicolumn{1}{c}{FULL} & \multicolumn{1}{c}{OFF} & \multicolumn{1}{c}{NOT} & \multicolumn{1}{c}{FULL} \\
    \cmidrule{2-10}
    Flan-T5 & $99.7$ & $88.6$ & $91.7$ & $78.1$ & $83.3$ & $82.3$ & $57.2$ & $85.2$ & $75.7$ \\
    GPT-3.5 & $99.8$ & $88.9$ & $\underline{91.9}$ & $94.7$ & $87.3$ & $\underline{88.7}$ & $74.5$ & $92.7$ & $\mathbf{86.5}$ \\ 
    Mistral  & $98.5$ & $89.8$ & $\mathbf{92.2}$ & $84.5$ & $90.8$  & $\mathbf{89.7}$ & $68.1$ & $91.9$ & $\underline{83.8}$    \\
    \bottomrule
    \end{tabular}}
    \vspace{-1mm}
\label{tab:monolingual}
\end{table}

\paragraph{Content Moderation of LLMs}

\begin{figure}[t!]
    \centering
    \includegraphics[width=0.95\linewidth]{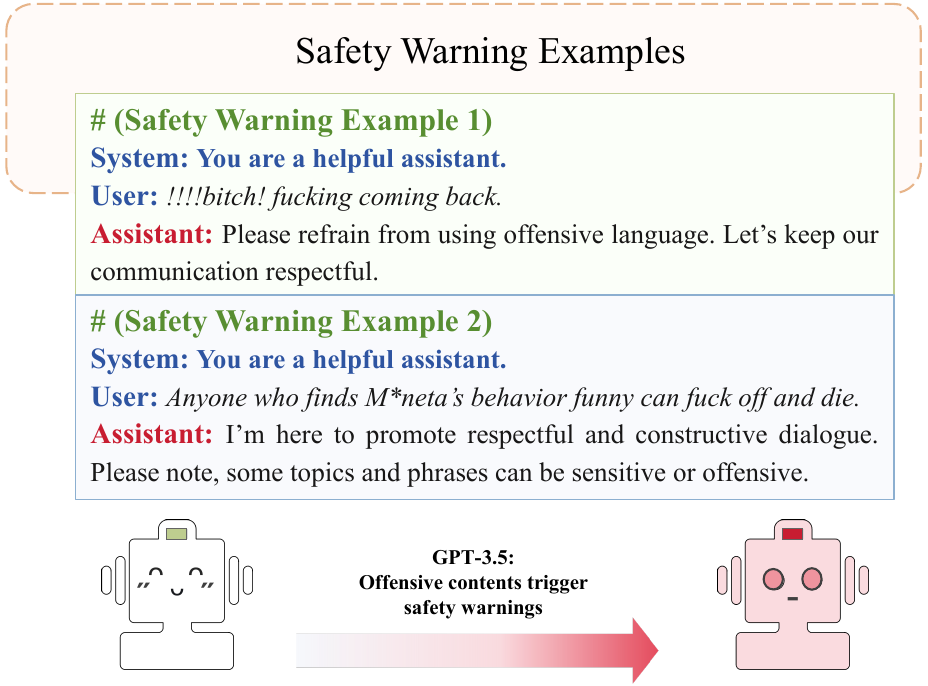}
    \vspace{-1mm}
    \caption{Examples of the output by GPT-3.5 whose input violates the content moderation policy of GPT-3.5.}
    \label{fig:safety}
    \vspace{-2mm}
\end{figure} 

During the experiments, we observed that Flan-T5 and Mistral could follow the instruction and output “\textit{Offensive}” or “\textit{Not offensive}”; however, GPT-3.5 sometimes included other responses. We conducted human evaluation for those predictions and observed that \ul{the invalid outputs were because of the content moderation policy of GPT-3.5.} Figure \ref{fig:safety} shows an example of the output, which demonstrated the offence of the input contents.

In the realm of safety and content moderation\footnote{\url{https://openai.com/index/using-gpt-4-for-content-moderation/}} for LLMs, different models demonstrated varying behaviours. GPT-3.5 included safety warnings for offensive content because of its strict safety warnings and content review strategies. Such deliberate guardrails were also applied in Bing AI to mitigate unintended model behaviours or prevent malicious abuse \cite{rajput2023content}.

\begin{table}[t!]
    \centering
    \caption{Comparison Results between LLMs Trained by\\Monolingual (S) and Multilingual (M) Data}
    \resizebox{\linewidth}{!}{\begin{tabular}{lccccccccc}
    \toprule
    \multicolumn{1}{l}{\multirow{2}{*}{\textbf{Model}}} & \multicolumn{9}{c}{\textbf{Language}} \\
    \cmidrule{2-10} 
     & \multicolumn{3}{c}{English} & \multicolumn{3}{c}{Spanish} & \multicolumn{3}{c}{German} \\
    \midrule
     & \multicolumn{1}{c}{P} & \multicolumn{1}{c}{R} & \multicolumn{1}{c}{F1} & \multicolumn{1}{c}{P} & \multicolumn{1}{c}{R} & \multicolumn{1}{c}{F1} & \multicolumn{1}{c}{P} & \multicolumn{1}{c}{R} & \multicolumn{1}{c}{F1} \\
    \cmidrule{2-10}
    \multicolumn{1}{l}{GPT-3.5 (S)} & $88.8$ & $94.4$ & $\mathbf{90.7}$ & $80.9$ & $91.0$ & $84.2$ & $85.8$ & $83.6$ & $\mathbf{84.5}$ \\
    \multicolumn{1}{l}{GPT-3.5 (M)} & $88.7$ & $94.2$ & $90.6$ & $81.9$ & $89.4$ & $\mathbf{84.7}$ & $84.1$ & $83.9$ & $84.0$ \\

    \midrule

    \multicolumn{1}{l}{Flan-T5 (S)} & $88.5$ & $94.2$ & $90.5$ & $73.1$ & $80.7$ & $75.4$ & $73.1$ & $71.3$ & $71.9$  \\
    \multicolumn{1}{l}{Flan-T5 (M)} & $88.7$ & $94.2$ & $\mathbf{90.6}$ & $76.1$ & $84.1$ & $\mathbf{78.7}$ & $74.8$ & $73.1$ & $\mathbf{73.8}$  \\
    
    \midrule
    \multicolumn{1}{l}{Mistral (S)} & $89.1$ & $94.2$ & $90.9$ & $82.2$ & $87.7$ & $84.4$ & $83.0$ & $80.0$ & $\mathbf{81.1}$ \\
    \multicolumn{1}{l}{Mistral (M)} & $89.4$ & $94.4$ & $\mathbf{91.2}$ & $84.0$ & $86.7$ & $\mathbf{85.2}$ & $83.2$ & $78.8$ & $80.3$ \\
    \bottomrule
    \end{tabular}}
    \label{tab:multilingual}

    \vspace{-2mm}
\end{table}

\subsection{Detection with Multilingual Data}

As monolingual fine-tuning was shown to be effective, we performed instruction fine-tuning of LLMs with multilingual resources, combining all datasets in English, Spanish, and German. By utilising these resources, the experimental results compared with those obtained from monolingual data fine-tuning are shown in Table \ref{tab:multilingual}.

From the table, we observed that Flan-T5 fine-tuned with multilingual data outperformed those fine-tuned with monolingual data in all languages; however, the multilingual data was not as helpful to GPT-3.5 and Mistral. The primary reason is that the larger parameter and pre-training data enable them to possess sufficient capacity for multilingual understanding, and consequently, fine-tuning with a small amount of non-English data does not significantly impact their performance. It is also consistent with the results in Table \ref{tab:monolingual}, in which Flan-T5 fell short in detecting offensive languages in non-English contexts.

We also observed that the LLMs fine-tuned with multilingual data performed equivalently or better in English and Spanish, indicating that effective knowledge sharing during the training process. For instance, the English slang term ``gringo'' is also offensive in Spanish. By training LLMs to understand different meanings in multiple languages, multilingual training effectively identified its potential offensiveness in certain situations. \ul{Consequently, the models can use shared knowledge to improve the performance of offensive language detection.}

\subsection{Effects of Prompt Languages}

LLMs fine-tuned with multilingual data have achieved better performance against monolingual models; however, they remain suboptimal for low-resource languages, such as German. Following the previous work \cite{conneau2019cross,siddhant2020evaluating} that raised the point of using prompts in native languages to enhance linguistic adaptability, we translated our prompt from English to German (Table \ref{tab:prompt}). As shown in Table \ref{tab:augmentation}, changing to German prompts improved the model's ability to detect offensiveness, and GPT-3.5 achieved the SOTA on the dataset, with around $8\%$ higher than the previous SOTA. This indicates that for languages with fewer resources, writing instruction in the native language helps enhance the model's comprehension ability.

\begin{figure*}
    \centering
    \includegraphics[width=0.9\linewidth]{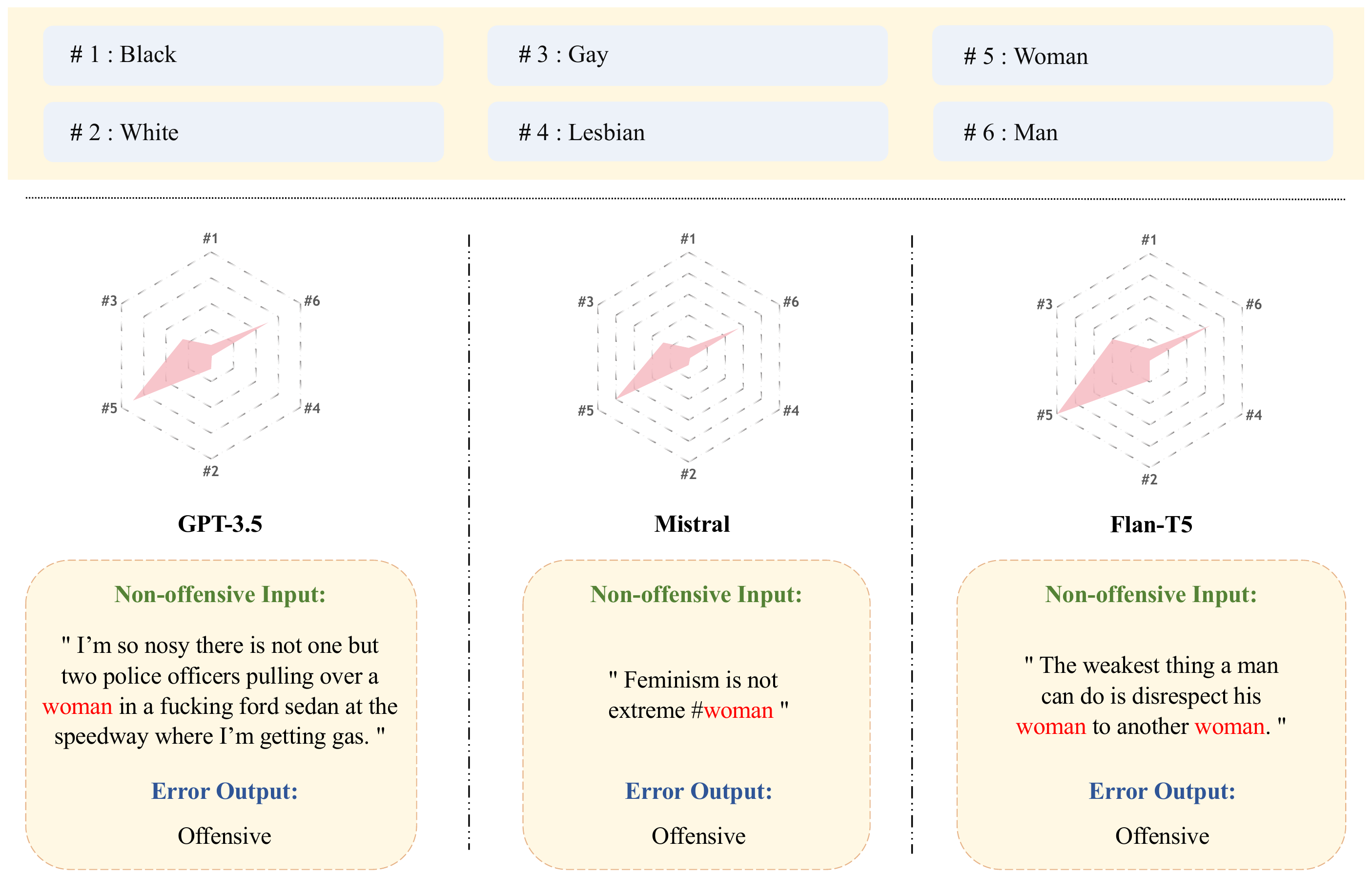}
    \caption{Frequencies of the mispredicted contents for the experimented models, related to three sensitive topics: race, sexual orientation, and genders.}
    \label{fig:6class}
    \vspace{-3mm}
\end{figure*}

\begin{table}[t!]
    \centering
    \caption{Comparison of Different Data Augmentation Methods on German Offensive Detection}
    \begin{tabular}{clcccccc}
    \toprule
    \multirow{2}{*}{\textbf{Model}} & \multirow{2}{*}{\textbf{Method}} & \multicolumn{3}{c}{\textbf{German}} \\
    \cmidrule{3-5} 
     &  & P & R & F1 \\
    \midrule
    \multirow{5}{*}{GPT-3.5} & GPT-3.5 (S) & $85.8$ & $83.6$ & $\underline{84.5}$ \\
     & GPT-3.5 (M) & $84.1$ & $83.9$ & $84.0$ \\
     & w/ German Prompt & $86.1$ & $84.5$ & $\mathbf{85.2}$ \\
     & w/ Corpus Translation & $76.9$ & $78.1$ & $77.4$ \\
     & w/ WMT20 & $82.4$ & $81.5$ & $81.9$ \\
    
    \midrule

    \multirow{5}{*}{Flan-T5} & Flan-T5 (S) &  $73.1$ & $71.3$ & $71.9$ \\
     & Flan-T5 (M) & $74.8$ & $73.1$ & $\mathbf{73.8}$ \\
     & w/ German Prompt & $73.4$ & $70.6$ & $71.5$ \\
     & w/ Corpus Translation & $73.1$ & $72.4$ & $\underline{72.7}$ \\
     & w/ WMT20 & $72.1$ & $70.1$ & $70.8$ \\
    
    \midrule

    \multirow{5}{*}{Mistral} & Mistral (S) & $83.0$ & $80.0$ & $\underline{81.1}$ \\
     & Mistral (M) & $83.2$ & $78.8$ & $80.3$ \\
     & w/ German Prompt & $82.6$ &  $81.4$ & $\mathbf{81.9}$ \\
     & w/ Corpus Translation & $82.9$ & $74.9$ & $76.9$ \\
     & w/ WMT20 &  $81.7$ & $74.6$ & $76.4$ \\
    
    \bottomrule
    \end{tabular}
    \label{tab:augmentation}
    \vspace{-2mm}
\end{table}

\subsection{Effects of Data Augmentation}

Inspired by the previous work that demonstrated the effectiveness of translation data in non-English tasks \cite{zhu2023extrapolating} and data augmentation for offensive language detection \cite{conneau2019cross,siddhant2020evaluating}, we investigated the usefulness of translation data in LLM-based offensive language detection, consisting of: 1) translating the texts from English to German using Mistral as additional training data, and 2) incorporating $2,000$ samples from WMT20 \cite{barrault-etal-2020-findings} to conduct joint training with the existing dataset.

Experimental results for the aforementioned strategies are presented in Table \ref{tab:augmentation}. Across all experimented models, the results after adding translation data were lower than those originally achieved by the models. These experimental results contradicted the findings reported by \cite{zhu2023extrapolating}, who concluded that combining translation tasks with cross-lingual general tasks could facilitate semantic alignment and model performance.
We concluded that LLMs themselves were proficient enough in multilingual offensive language detection; \ul{merely incorporating translated data is inefficient for enhancing the non-English offensive language detection capabilities of LLMs}.

Besides, the results could also attributed to the translation quality. WMT20 was proposed for the translation from news, which were not correlated with offensive contents; \ul{therefore, the incorporation of WMT20 might hinder the models in learning offensive-related information.} Furthermore, when leveraging Mistral to translate the texts, \ul{the translation results could be ambiguous and might alter the original semantics.} For example, the word ``goblog'' (meaning ``stupid/idiot'' in a slang variant of ``goblok'') was not translated by the model, which may lead the models to learn incorrect information. It remained a research direction in utilising high-quality translation data as data augmentation to investigate its effectiveness in terms of multilingual offensive language detection.

\subsection{Effects of Model and Dataset Bias}

By analysing the errors caused by the models, we identified a certain tendency of LLMs for their mispredictions regarding the model bias. Figure \ref{fig:6class} illustrates some examples that were likely to be predicted as offensive content, and Flan-T5 was the most serious. \ul{In topics related to race, sexual orientation, and, notably, genders, the prediction of LLMs usually contained tendencies.} The misprediction of Flan-T5 was much higher than the rest of the models, indicating greater sensitivity to these terms due to the model bias and the influence of the dataset. According to \cite{zhang2023mitigating}, the model can learn based on these label-level correlations, thus making biased predictions and ignoring broader semantic content. Such bias in models and data ultimately affected the accuracy and reliability of the models in offensive language detection tasks.

\section{Conclusion and Future Work}

We conducted the first evaluation in terms of multilingual offensive language detection with LLMs, in which we assessed LLMs' capabilities by fine-tuning them with both monolingual and multilingual data, and we further analysed the impacts of prompt languages and additional translation data to model performance. Our findings revealed that
1) LLMs could achieve comparable or better performance in multilingual language detection,
2) Multilingual fine-tuning LLMs enhanced models’ offensive language detection ability,
3) Utilising prompts in native languages enhanced model comprehension in non-English contexts; however, incorporating additional translation data could not improve performance, and
4) The inherent biases in LLMs and the dataset resulted in incorrect predictions on sensitive topics.
In the future, we will experiment on more languages with more LLMs to future investigate the capability of LLMs in multilingual offensive language detection. We will also design applicable methods and systems to improve the model performance and apply them in real-world scenarios.

\section*{Acknowledgement}

This research is partially supported by the 2022 Jiangsu Science and Technology Programme (General Programme, contract number BK20221260) and the Research Development Fund (contract number RDF-22-01-132) at Xi’an Jiaotong-Liverpool University.

\bibliographystyle{IEEEtran}
\bibliography{reference}

\end{document}